\documentclass[12pt]{article}
\usepackage[dvips]{color,graphicx}
\usepackage{amsmath,amssymb}
\usepackage{epsf}
\usepackage{natbib}

\newtheorem{theorem}{Theorem}
\newtheorem{lemma}[theorem]{Lemma}
\newtheorem{corollary}[theorem]{Corollary}

\newtheorem{definition}[theorem]{Definition}

\title{Asymptotic Accuracy of Bayesian Estimation\\
for a Single Latent Variable}

\author{Keisuke Yamazaki\\
       k-yam@math.dis.titech.ac.jp \\
       Department of Computational Intelligence and Systems Science,\\
       Tokyo Institute of Technology\\
       G5-19 4259 Nagatsuta, Midori-ku, Yokohama, Japan
	}
\date{}
\begin{document}
\maketitle

\begin{abstract}
In data science and machine learning,
hierarchical parametric models, such as mixture models, are often used.
They contain two kinds of variables: observable variables, 
which represent the parts of the data that can be directly measured, and latent variables, 
which represent the underlying processes that generate the data.
Although there has been an increase in research on the estimation accuracy for observable variables,
the theoretical analysis of estimating latent variables has not been thoroughly investigated. 
In a previous study, we determined the accuracy of a Bayes estimation
for the joint probability of the latent variables in a dataset, and we
proved that the Bayes method is asymptotically more accurate than the maximum-likelihood method.
However, the accuracy of the Bayes estimation for a single latent variable remains unknown.
In the present paper,
we derive the asymptotic expansions of the error functions, which are defined
by the Kullback-Leibler divergence, for two types of single-variable estimations
when the statistical regularity is satisfied.
Our results indicate that the accuracies of the Bayes and maximum-likelihood methods are asymptotically equivalent
and clarify that the Bayes method is only advantageous for multivariable estimations.
\end{abstract}

\section{Introduction}
\sloppy
In machine learning and data science,
hierarchical parametric models, such as mixture models, are often used.
These models contain two kinds of variables: observable  and latent.
The observable variables represent the observable, measurable data,
while the latent variables express the underlying processes that generate the data.
For example, a common hierarchical model is a mixture of Gaussian distributions defined by
\begin{align*}
p(x|w) =& \sum_{k=1}^K a_k\mathcal{N}(x|\mu_k,\Sigma),
\end{align*}
where $x\in R^M$ is the observable position,
$w$ is the parameter containing $a_k$ and $\mu_k$,
$a_k\ge 0$ is the mixing ratio,
and $\mathcal{N}(x|\mu,\Sigma)$ is a Gaussian distribution with mean $\mu$
and variance-covariance matrix $\Sigma$.
Let us consider cluster analysis, 
which is a typical task of unsupervised learning.
The observable variable is the data position $x$,
and the latent variable is the ungiven cluster label $k \in \{1,\dots,K\}$, which indicates to which component/cluster
the data belong.

Since the parameter is unknown, in practice, it is often 
necessary to deal with both the parameter and the observable or the latent variable.
The parameter is usually estimated in one of two ways:
the maximum-likelihood method or the Bayes method.
The maximum-likelihood method estimates the parameter that 
maximizes the likelihood function, 
while the Bayes method determines the optimal (posterior) distribution for the parameter.

It has been noted that the hierarchical models include
singularities in the parameter space \citep{Amari,Watanabe01b}.
At a singular point, the relation
between the parameter $w$ and the probability $p(x|w)$ is not one to one,
and the Fisher information matrix is not positive definite.
Let the $K^*$ component Gaussian mixture be the data-generating distribution,
and let the $K$ component mixture be a learning model.
The case $K>K^*$ corresponds to a singular case: 
there are redundant components and their parameters contain singularities.
On the other hand, the well-specified case $K=K^*$ does not have singularities,
and in the present paper, we call it a regular case.

The estimation of an unseen observable variable is referred to as a prediction.
Let a set of the given data be $X^n=\{x_1,\dots,x_n\}$.
The task is to predict the next data position based on the given data;
this is formulated as the estimation of the probability $p(x_{n+1}|X^n)$.
In order to measure the accuracy of the task,
we define the error function to be the Kullback-Leibler divergence,
\begin{align*}
E_{X^n}\bigg[ \int q(x_{n+1})\ln \frac{q(x_{n+1})}{p(x_{n+1}|X^n)}dx_{n+1}\bigg],
\end{align*}
where $q(x)$ is the data-generating distribution
and $E_{X^n}[\cdot]$ is the expectation over all of the given data.
In the example of the Gaussian mixture,
the prediction task is to estimate the next unseen data positions.

The estimation of the latent variables is not the same as the prediction task.
The target variable of the estimation is unobservable,
and in many practical situations, its true value is not given;
this makes it difficult to evaluate the result.
In a previous study \citep{Yamazaki14a},
we formulated the accuracy of the latent-variable estimation
in a distribution-based manner.
The estimation of latent variables is divided into three classes.
Let a set of latent variables be $Y^n=\{y_1,\dots,y_n\}$,
where $y_i$ is the corresponding variable to $x_i$.
\begin{figure}[t]
\centering
\includegraphics[angle=-90,width=0.7\columnwidth]{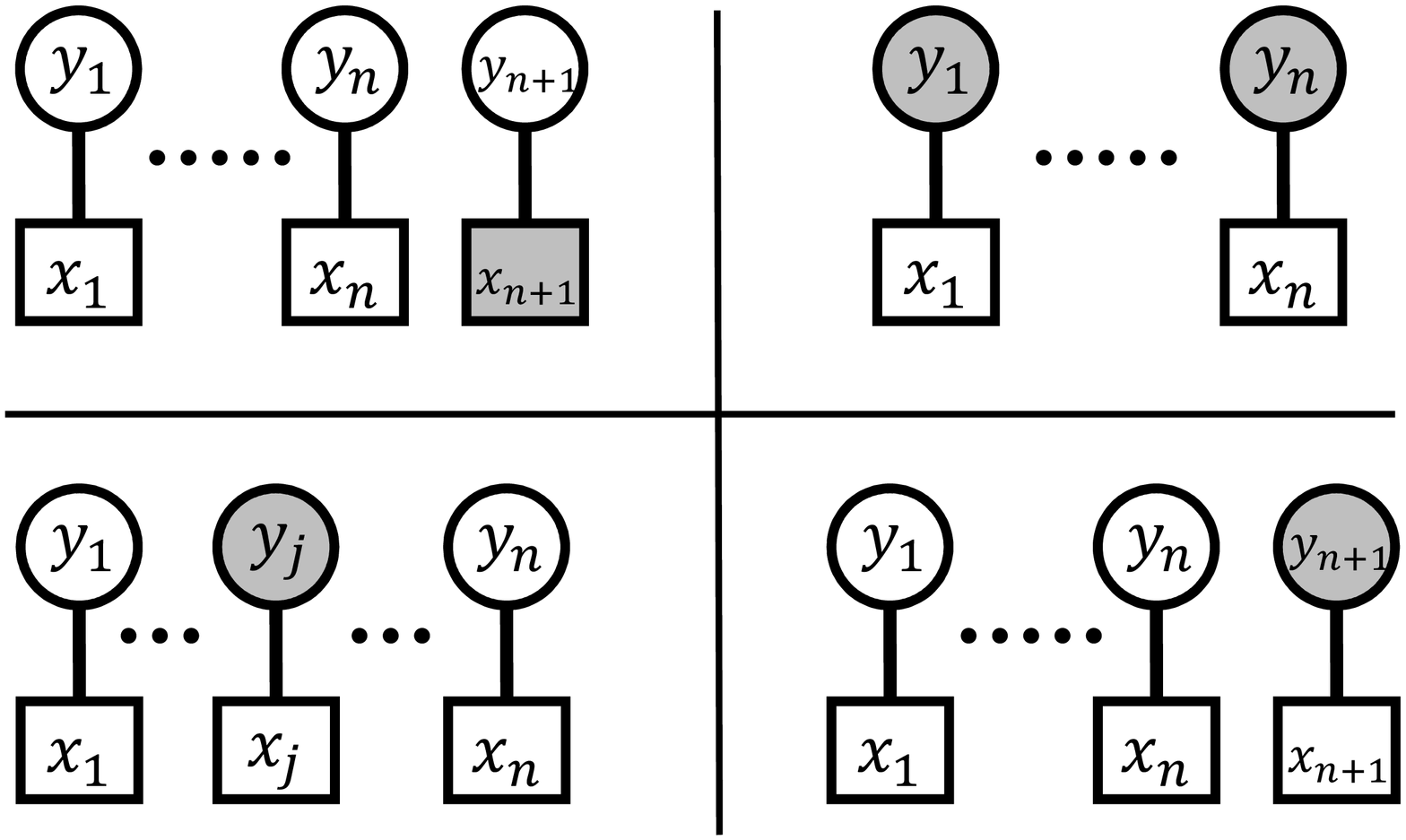}
\caption{Predictions of observable variables and estimations of latent variables.
The observable data are $\{x_1,\dots,x_n\}$. Rectangles and circles represent the observable and unobservable variables, respectively.
Gray nodes are the estimation targets.}
\label{fig:estimations}
\end{figure}
Figure \ref{fig:estimations} shows the prediction of observable variables
and the three types of estimations of latent variables.
Rectangles and circles indicate the observable and latent variables, respectively.
The gray nodes are the targets of the estimations.
The top left panel shows the prediction,
which is expressed as the estimation of $p(x_{n+1}|X^n)$.
The top right panel shows the estimation of the joint probability $p(Y^n|X^n)$,
in which 
all of the latent variables are targets; we will refer to this as Type I.
The bottom left panel shows the estimation of the probability 
of a specific latent variable $p(y_j|X^n)$;
we will refer to this as Type II.
The bottom right panel shows the estimation of the probability 
of a latent variable in the unseen data $p(y_{n+1}|X^n)$;
we will refer to this as Type III.
In the example of a Gaussian mixture,
these three types of latent-variable estimation correspond to the cluster analysis process of
assigning labels to data.

When the number of data points $n$ is sufficiently large,
the form of the error function is referred to as the asymptotic expansion,
and the calculation of this form has been exhaustively studied for the prediction process.
In the maximum-likelihood method,
the asymptotic error is well known, and it has been used
as a criterion for selecting models \citep{Akaike,Takeuchi76,Econometrica:White:1982}.
In the Bayes method,
the estimation depends on the posterior distribution,
and the theoretical properties of its convergence have been studied 
\citep{LeCam1973,Ghosal+2000,Nguyen2013}.
The normalizing factor of the posterior distribution is the marginal likelihood,
and this has a direct relation with the error function \citep{Levin}.
Since the asymptotic expansion of the marginal likelihood has been derived
for the regular case \citep{Schwarz,Clarke90},
this relation allows us to calculate the asymptotic error.
In the singular case, algebraic geometry plays an effective role; in particular,
the resolution of singularities \citep{Hironaka} can be used to
clarify the asymptotic marginal likelihood and the asymptotic error
\citep{Watanabe01a,Aoyagi04,Yamazaki03a,Rusakov,Watanabe09:book,Zwiernik11,Naito14}.

These studies on the predictive error have focused on the estimation
of a single observable variable.
Based on their definitions, in the maximum-likelihood method, the error function
for the joint probability of multiple variables
is equivalent to that for the probability of a single variable. 
The form of the error of the Bayes method depends on the number of variables.
For the regular case, 
an information criterion that uses the asymptotic error of the joint probability was devised for use with the selection of a Bayesian model \citep{Ando2007}.

Although there are a number of studies 
that consider the estimation of observable variables and the convergence of the parameters,
the theory of estimating latent variables has not been thoroughly analyzed.
The error functions of Types I, II, and III are defined as the Kullback-Leibler divergence
from the data-generating distribution to the estimated one,
and its theoretical behavior has been analyzed. 
The error function of Type III with the maximum-likelihood method has been derived,
and a model-selection criterion has been proposed for the regular case \citep{ShimodairaPDOI}.
The asymptotic expansions of Type I in the Bayes method
and of the rest of the types in the maximum-likelihood method have been calculated for the regular case,
and we found that 
with the maximum-likelihood method, their asymptotic errors are equivalent
and that for Type I, the Bayes method is more accurate than the maximum-likelihood method.
The singular case has been considered, and its error has been derived only for Type I \citep{Yamazaki15a}.

The asymptotic errors of Types II and III with the Bayes method are as yet unknown
in both the regular and the singular cases.
Since the asymptotic analysis for these estimations of a single variable requires
the calculation of higher-order terms of the marginal likelihood, deriving the asymptotic expansions
is not straightforward.
In the present paper,
we reveal one of the higher-order terms and show the asymptotic errors of Types II and III
for the regular case.
Comparing the results of this to those of the maximum-likelihood method, 
we determined that the Bayes method is advantageous only for multivariable estimations, such as those for Type I.

The remainder of this paper is organized as follows:
The three types of estimations and their error functions are formally defined in Section \ref{sec:formulation}.
The results from our previous study are introduced in Section \ref{sec:known}.
Section \ref{sec:results} presents the main results on the accuracy of estimations of Types II and III.
The advantage of the Bayes estimation is discussed in Section \ref{sec:dis}.
\section{Three Types of Estimations of Latent Variables}
\label{sec:formulation}
In this section, we formulate the three types of estimations of latent variables.
\subsection{Formulation of a Hierarchical Probabilistic Model}
Let $x \in R^M$ and $y\in \{1,\dots,K\}$ be observable and latent variables, respectively.
The model is represented by
\begin{align*}
p(x,y|w) =& p(y|w)p(x|y,w),
\end{align*}
where the parameter is expressed as $w \in W \subset R^d$.
The probabilistic density function of $x$ is then expressed as
\begin{align*}
p(x|w) =& \sum_{y=1}^K p(x,y|w) = \sum_{y=1}^K p(y|w)p(x|y,w).
\end{align*}
In the data-generating process of the rightmost expression, 
we assume that $y$ is selected based on $p(y|w)$,
and then $x$ is determined by $p(x|y,w)$.
In machine learning, this mixture-type form is used to express many hierarchical models,
such as Bayesian networks.

Let $\{X^n,Y^n\}=\{(x_1,y_1),\dots,(x_n,y_n)\}$ be the i.i.d. data set
generated by the true distribution $q(x,y)$.
We assume that the true distribution is expressed as
\begin{align*}
q(x,y) =& p(x,y|w^*),
\end{align*}
where $w^*$ is referred to as the true parameter.
\subsection{The Three Estimations and their Error Functions}
First, we introduce the maximum-likelihood estimator and the posterior distribution,
which play important roles in the maximum-likelihood and Bayes methods, respectively.
The likelihood function is defined by
\begin{align*}
L(w) =& \prod_{i=1}^n p(x_i|w).
\end{align*}
The maximum-likelihood estimator is given by
\begin{align*}
\hat{w} =& \arg\max_w L(w).
\end{align*}
In the maximum-likelihood method,
this estimator is regarded as the optimal parameter.
For example, the prediction of unseen data $x$ is given by
\begin{align*}
p(x|X^n) = p(x|\hat{w}).
\end{align*}

On the other hand, the Bayes method is defined
based on the posterior distribution $p(w|X^n)$.
Using a prior distribution $\varphi(w)$,
we define the posterior distribution as
\begin{align*}
p(w|X^n) =& \frac{1}{Z(X^n)}L(w)\varphi(w),
\end{align*}
where $Z(X^n)$ is a normalizing factor given by
\begin{align*}
Z(X^n) =& \int L(w)\varphi(w)dw = \int \prod_{i=1}^n p(x_i|w)\varphi(w)dw.
\end{align*}
The prediction $p(x|X^n)$ is given by
\begin{align*}
p(x|X^n) = \int p(x|w)p(w|X^n)dw.
\end{align*}
We assume that the true parameter $w^*$ is included in the support of the prior distribution.

Next, for both the maximum-likelihood and Bayes methods,
we define the estimated probabilities of the latent variable for each of the three types.
For Type I, the given data are $X^n=\{x_1,\dots,x_n\}$,
and the estimation target is $Y^n=\{y_1,\dots,y_n\}$.
The estimated probability of the maximum-likelihood estimation is defined by
\begin{align*}
p(Y^n|X^n) =& \prod_{i=1}^n p(y_i|x_i,\hat{w}) = \prod_{i=1}^n \frac{p(x_i,y_i|\hat{w})}{p(x_i|\hat{w})}.
\end{align*}
In the Bayes estimation, it is defined by
\begin{align*}
p(Y^n|X^n) =& \int \prod_{i=1}^n p(y_i|x_i,w)p(w|X^n)dw\\
=& \int \prod_{i=1}^n\frac{p(x_i,y_i|w)}{p(x_i|w)}p(w|X^n)dw\\
=& \frac{\int \prod_{i=1}^n p(x_i,y_i|w)\varphi(w)dw}{\int \prod_{i=1}^n p(x_i|w)\varphi(w)dw}.
\end{align*}
For Type II, the given data are $X^n$, and the estimation target is one of the elements in $Y^n$.
Let the target be $y_j\in Y^n$.
The estimated probability of the maximum-likelihood estimation is defined by
\begin{align*}
p(y_j|X^n) =& p(y_j|x_j,\hat{w}) = \frac{p(x_j,y_j|\hat{w})}{p(x_j|\hat{w})}.
\end{align*}
In the Bayes estimation, it is defined by
\begin{align*}
p(y_j|X^n) =& \int p(y_j|x_j,w)p(w|X^n)dw\\
=& \int \frac{p(x_j,y_j|w)}{p(x_j|w)}p(w|X^n)dw\\
=& \frac{\int p(x_j,y_j|w)\prod_{i\ne j}p(x_i|w)\varphi(w)dw}{\int \prod_{i=1}^n p(x_i|w)\varphi(w)dw}.
\end{align*}
For Type III, the given data are $X^{n+1}=\{X^n,x_{n+1}\}$, and the estimation target is $y_{n+1}$.
The estimated probability of the maximum-likelihood estimation is defined by
\begin{align*}
p(y_{n+1}|X^{n+1}) =& p(y_{n+1}|x_{n+1},\hat{w}) = \frac{p(x_{n+1},y_{n+1}|\hat{w})}{p(x_{n+1}|\hat{w})}.
\end{align*}
In the Bayes estimation, it is defined by
\begin{align*}
p(y_{n+1}|X^{n+1}) =& \int p(y_{n+1}|x_{n+1},w)p(w|X^n)dw\\
=& \frac{\int p(y_{n+1}|x_{n+1},w)\prod_{i=1}^n p(x_i|w)\varphi(w)dw}{\int \prod_{i=1}^n p(x_i|w)\varphi(w)dw}.
\end{align*}

Finally, we define the error functions that measure the accuracy of these estimations, and these are
based on the average Kullback-Leibler divergence.
In Type I, the true probability of $Y^n$ is expressed by
\begin{align*}
q(Y^n|X^n) =& \prod_{i=1}^n q(y_i|x_i) = \prod_{i=1}^n \frac{q(x_i,y_i)}{q(x_i)}.
\end{align*}
The error function is given by
\begin{align*}
D_{\mathrm{I}}(n) =& \frac{1}{n} E_n\bigg[ \ln \frac{q(Y^n|X^n)}{p(Y^n|X^n)}\bigg],
\end{align*}
where the expectation is described as
\begin{align*}
E_n[f(X^n,Y^n)] =& \int \sum_{y_1=1}^K \dots \sum_{y_n=1}^K q(X^n,Y^n) f(X^n,Y^n) dx_1\dots dx_n.
\end{align*}
In Types II and III, the error functions are given by
\begin{align*}
D_{\mathrm{II}}(n) =& \frac{1}{n}\sum_{j=1}^n E_n\bigg[ \ln \frac{q(y_j|x_j)}{p(y_j|X^n)}\bigg],\\
D_{\mathrm{III}}(n) =& E_{n+1} \bigg[ \ln \frac{q(y_{n+1}|x_{n+1})}{p(y_{n+1}|X^{n+1})}\bigg],
\end{align*}
respectively.
\section{Previous Results on Asymptotic Error Functions}
\label{sec:known}
This section presents results that we published previously \citep{Yamazaki14a}.
We obtained the asymptotic expansions of $D_{\mathrm{I}}(n)$
for both estimation methods,
and the asymptotic expansions of $D_{\mathrm{II}}(n)$ and $D_{\mathrm{III}}(n)$ for the maximum-likelihood estimation.
The Fisher information matrices of $p(x,y|w)$ and $p(x|w)$ are defined as
\begin{align*}
\{I_{XY}(w)\}_{ij} =& \int \sum_{y=1}^K \frac{\partial \ln p(x,y|w)}{\partial w_i}\frac{\partial \ln p(x,y|w)}{\partial w_j} p(x,y|w)dx,\\
\{I_X(w)\}_{ij} =& \int \frac{\partial \ln p(x|w)}{\partial w_i}\frac{\partial \ln p(x|w)}{\partial w_j} p(x|w)dx,
\end{align*}
respectively.
Let $I_{Y|X}(w)$ be their difference:
\begin{align*}
I_{Y|X}(w) =& I_{XY}(w) - I_X(w).
\end{align*}
In the present paper, we assume that these Fisher information matrices exist
and that the maximum-likelihood estimator converges almost surely to $w^*$ \citep{Wald1949}.
In other words, the models $p(x,y|w)$ and $p(x|w)$ are regular around $w^*$,
and the estimator is consistent \citep{vdVaart98}.

Because the latent variable is not observable,
there is a set of symmetric points $W^*_X$ 
such that $q(x)=p(x|w^*_X)$ for $w^*_X\in W^*_X$.
Note that the true parameter $w^*$ is one of the elements of $W^*_X$.
For example, let us consider the two-component Gaussian mixture given by
\begin{align*}
p(x|w) = a \mathcal{N}(x|\mu_1,\Sigma) + (1-a)\mathcal{N}(x|\mu_2,\Sigma),
\end{align*}
and let us assume that the true distribution is defined by
\begin{align*}
q(x,y=1) =& a^* \mathcal{N}(x|\mu^*_1,\Sigma),\\
q(x,y=2) =& (1-a^*)\mathcal{N}(x|\mu^*_2,\Sigma),
\end{align*}
where $a^*$, $\mu^*_1$, and $\mu^*_2$ are constants.
This means that the true parameter $w^*$ is described by
\begin{align*}
w^* =& (a^*,\mu_1^*,\mu_2^*)^\top.
\end{align*}
The parameter $w^*_s$ given by
\begin{align*}
w^*_s =& (1-a^*,\mu^*_2,\mu^*_1)^\top
\end{align*}
also satisfies $q(x)=p(x|w^*_s)$, where the labels $y=1,2$ are switched.
Then, $W^*_X=\{w^*,w^*_s\}$, and we refer to these points as symmetric,
since they provide symmetric label assignments.
This symmetry appears when $K\ge 2$.

Thus, the maximum-likelihood estimator does not always converge to $w^*$, and
in cluster analysis, this is known as the label-switching problem.
To avoid this problem and to theoretically analyze the error function,
we consider the case $\hat{w}\rightarrow w^*$.

Under the above assumptions, the following theorem has been proven.
\begin{theorem}
\label{th:prev_results}
The error functions have the following asymptotic expansion:
\begin{align*}
D(n) =& \frac{c}{n} + o\bigg(\frac{1}{n}\bigg),
\end{align*}
where $D(n)$ is a general notation for $D_{\mathrm{I}}(n)$, $D_{\mathrm{II}}(n)$,
and $D_{\mathrm{III}}(n)$, and the coefficient $c$ for each case
is shown in Table \ref{tab:errors}.
\begin{table}[t]
\centering
\caption{Coefficients of the dominant order $1/n$ in the error functions}
\begin{tabular}{|c|c|c|c|}
\hline
& Type I & Type II & Type III \\
\hline
ML & $\mathrm{Tr}[I_{Y|X}I_X^{-1}]/2$ & $\mathrm{Tr}[I_{Y|X}I_X^{-1}]/2$ & $\mathrm{Tr}[I_{Y|X}I_X^{-1}]/2$ \\
\hline
Bayes & $\ln\det[I_{XY}I_X^{-1}]/2$ & unknown & unknown\\
\hline
\end{tabular}
\label{tab:errors}
\end{table}
%
The rows indicate the maximum-likelihood (ML) and Bayes methods, respectively.
The matrices $I_{XY}(w^*)$, $I_X(w^*)$, and $I_{Y|X}(w^*)$ are abbreviated
in a form that does not include the true parameter, i.e., $I_{XY}$, $I_X$, or $I_{Y|X}$, respectively.
\end{theorem}
The following corollary compares the two estimation methods with Type I,
and shows the advantages of the Bayes estimation.
\begin{corollary}
\label{cor:comp}
Let the error functions for the maximum-likelihood and Bayes methods be denoted by
$D^{\mathrm{ML}}_{\mathrm{I}}(n)$ and $D^{\mathrm{Bayes}}_{\mathrm{I}}(n)$, respectively.
For any true parameter $w^*$, there exists a positive constant $c_d$ such that
\begin{align*}
D^{\mathrm{ML}}_{\mathrm{I}}(n) - D^{\mathrm{Bayes}}_{\mathrm{I}}(n) \ge \frac{c_d}{n} + o\bigg(\frac{1}{n}\bigg).
\end{align*}
\end{corollary}
Corollary \ref{cor:comp} indicates that, based on the leading term in the error function,
$D^{\mathrm{ML}}_{\mathrm{I}}(n)>D^{\mathrm{Bayes}}_{\mathrm{I}}(n)$
in the asymptotic case of large $n$.
\section{Main Results}
\label{sec:results}
This section presents the asymptotic expansions of the error functions for Types II and III.
\subsection{Asymptotic Errors of Types II \& III in the Bayes Method}
Due to the assumptions about the Fisher information matrices 
and the convergence of the maximum-likelihood estimator,
we can determine that
\begin{align}
|\hat{w}-w^*|=O_p\bigg(\frac{1}{\sqrt{n}}\bigg).\label{eq:MLE_order}
\end{align}
Let $\hat{w}_{n-1}(j)$ be the maximum-likelihood estimator based on the dataset $X^n\setminus x_j$:
\begin{align*}
\hat{w}_{n-1}(j) =& \arg\max_w \prod_{i\ne j}^n p(x_i|w).
\end{align*}
In order to simplify the notation, we will use $\hat{w}_{n-1}$ for $\hat{w}_{n-1}(j)$.
This estimator also converges to the true parameter, and
\begin{align}
|\hat{w}_{n-1}-w^*| =& O_p\bigg(\frac{1}{\sqrt{n}}\bigg).\label{eq:MLE_n-1_order}
\end{align}

Now, we consider the asymptotic expansions of the error functions.
In the Bayes method, the error functions $D_{\mathrm{II}}(n)$
and $D_{\mathrm{III}}(n)$ are written as
\begin{align}
D_{\mathrm{II}}(n) =& 
\frac{1}{n}\sum_{j=1}^n\big\{ E_n\big[\ln q(x_j,y_j)-\ln q(x_j)\big]
+ F_1(n) - F_2(n)\big\},\label{eq:DIIFF}\\
D_{\mathrm{III}}(n) =& 
E_{n+1}\big[\ln q(y_{n+1}|x_{n+1})\big]+ F_3(n) - F_2(n),\label{eq:DIIIFF}
\end{align}
respectively, where
\begin{align*}
F_1(n) =& E_n\bigg[ - \ln \int p(x_j,y_j|w)\prod_{i\ne j}^n p(x_i|w)\varphi(w)dw \bigg],\\
F_2(n) =& E_n\bigg[ - \ln \int \prod_{i=1}^n p(x_i|w)\varphi(w)dw \bigg],\\
F_3(n) =& E_{n+1}\bigg[ - \ln \int p(y_{n+1}|x_{n+1},w)\prod_{i=1}^n p(x_i|w)\varphi(w)dw \bigg].
\end{align*}
Then, the asymptotic expansions of $F_1(n)$, $F_2(n)$, and $F_3(n)$ are necessary.
Let us define the following negative log marginal likelihood:
\begin{align*}
F_\xi(n) = & E_zE_n\bigg[ - \ln \int \xi(z|w)\prod_{i=1}^n p(x_i|w)\varphi(w)dw \bigg].
\end{align*}
This is a unified expression for $F_1(n)$, $F_2(n)$, and $F_3(n)$, in which
the function $\xi(z|w)$ will be replaced with
the parametric models $p(x_j,y_j|w)$, $p(x_i|w)$, and $p(y_{n+1}|x_{n+1},w)$, respectively.
Thus, we assume that $z$ is independent of $x_i$.
The expectation $E_z[\cdot]$ is based on $\xi(z|w^*)$.
In the case of $\xi(z|w)=p(x,y|w)$,
the expectation is defined as
\begin{align*}
E_z[f(z)] = & \int \sum_{y=1}^K f(x,y) p(x,y|w^*)dx,
\end{align*}
and the function $F_1(n)$ is given by
\begin{align*}
F_1(n) =& F_\xi(n-1).
\end{align*}

The following lemma plays an essential role 
in the asymptotic analysis of the error functions.
\begin{lemma}
\label{lem:Fxi}
The function $F_\xi(n)$ is expressed as
\begin{align}
F_\xi(n) =& -E_n\bigg[ \sum_{i=1}^n \ln p(x_i|\hat{w})\bigg] +\frac{d}{2}\ln n \nonumber\\
&- E_zE_n\bigg[ \ln \xi(z|\hat{w})\varphi(\hat{w})\bigg] \nonumber\\
&- \frac{1}{2}\ln 2\pi + \frac{1}{2}\ln \det I_X(w^*)\nonumber\\
&-\frac{1}{2n}\mathrm{Tr} E_n\bigg[ \frac{1}{\varphi(\hat{w})}
\frac{\partial^2 \varphi(\hat{w})}{\partial w^2}\bigg]I_X(w^*)^{-1} + o\bigg(\frac{1}{n}\bigg). \label{eq:Fxi}
\end{align}
\end{lemma}
This form contains terms of the order $1/n$,
and this order is higher than the constant terms derived in \citet{Clarke90}.
Note that $\xi(z|w)$ does not affect the $1/n$-order terms.
Since the error functions are the differences between the $F_\xi(n)$, as shown in Eqs. \ref{eq:DIIFF} and \ref{eq:DIIIFF},
it is easy to see that the error functions depend on
only the first and the third terms of Eq.~\ref{eq:Fxi},
which include the maximum-likelihood estimator $\hat{w}$.
This implies a connection between the Bayes method
and the maximum-likelihood method.

Based on this lemma,
we can prove the following two theorems.
\begin{theorem}
\label{th:DII}
Let the error functions for the maximum-likelihood and Bayes methods be denoted by
$D^{\mathrm{ML}}_{\mathrm{II}}(n)$ and $D^{\mathrm{Bayes}}_{\mathrm{II}}(n)$, respectively.
Asymptotically, they have the following relation:
\begin{align*}
D^{\mathrm{Bayes}}_{\mathrm{II}}(n) =& D^{\mathrm{ML}}_{\mathrm{II}}(n) + o\bigg(\frac{1}{n}\bigg)\\
=& \frac{\mathrm{Tr}I_{Y|X}(w^*)I_X(w^*)^{-1}}{2n} + o\bigg(\frac{1}{n}\bigg).
\end{align*}
\end{theorem}
\begin{theorem}
\label{th:DIII}
Let the error functions for the maximum-likelihood and Bayes methods be denoted by
$D^{\mathrm{ML}}_{\mathrm{III}}(n)$ and $D^{\mathrm{Bayes}}_{\mathrm{III}}(n)$, respectively.
Asymptotically, they have the following relation:
\begin{align*}
D^{\mathrm{Bayes}}_{\mathrm{III}}(n) =& D^{\mathrm{ML}}_{\mathrm{III}}(n) + o\bigg(\frac{1}{n}\bigg)\\
=& \frac{\mathrm{Tr}I_{Y|X}(w^*)I_X(w^*)^{-1}}{2n} + o\bigg(\frac{1}{n}\bigg).
\end{align*}
\end{theorem}
In Types II and III, the asymptotic errors of the Bayes estimation
are equivalent to those of the maximum-likelihood estimation.
Since Table \ref{tab:errors} shows 
$D^{\mathrm{ML}}_{\mathrm{II}}(n)=D^{\mathrm{ML}}_{\mathrm{III}}(n)$,
the errors of Types II and III are also asymptotically the same as those for the Bayes method.

The following corollary summarizes the relative magnitudes of the error functions.
\begin{corollary}
\label{cor:mag_rel}
Based on the leading terms of the error functions,
the relative magnitudes are as follows:
\begin{align*}
D^{\mathrm{Bayes}}_{\mathrm{I}}(n)< D^{\mathrm{ML}}_{\mathrm{I}}(n)
= D^{\mathrm{Bayes}}_{\mathrm{II}}(n) = D^{\mathrm{ML}}_{\mathrm{II}}(n)
= D^{\mathrm{Bayes}}_{\mathrm{III}}(n) = D^{\mathrm{ML}}_{\mathrm{III}}(n).
\end{align*}
\end{corollary}
Considering these results, in Section \ref{sec:dis}, we will discuss why the Bayes estimation
is more accurate than the maximum-likelihood estimation for the Type I estimation.
\subsection{Proof of Lemma \ref{lem:Fxi}}
Based on a saddle-point approximation, we have
\begin{align}
F_\xi(n) =& E_zE_n \bigg[ -\sum_{i=1}^n \ln p(x_i|\hat{w})
- \frac{1}{2} \ln 2\pi\det\{nI_X(w^*)\}^{-1} \nonumber\\
&-\ln \int g(w)\mathcal{N}(w|\hat{w}_n,\{nI_X(w^*)\}^{-1})dw\bigg],\label{eq:Fxi_mid}
\end{align}
where $g(w)=\xi(z|w)\varphi(w)e^{r(w)}$ and $r(w)=O_p((w-\hat{w})^3)$.
According to Eq. \ref{eq:MLE_order} and the asymptotic distribution of $\hat{w}$,
\begin{align}
&-\ln \int g(w)\mathcal{N}(w|\hat{w},\{nI_X(w^*)\}^{-1})dw\nonumber\\
&=-\ln g(\hat{w}) - \ln \int \bigg\{1+\frac{1}{2g(\hat{w})}(w-\hat{w})^\top 
\frac{\partial^2 g(\hat{w})}{\partial w^2}(w-\hat{w})+\dots\bigg\}\nonumber\\
&\;\;\;\; \times\mathcal{N}(w|\hat{w},\{nI_X(w^*)\}^{-1})dw\nonumber\\
&= -\ln g(\hat{w}) -\ln \bigg\{1+ \frac{1}{2g(\hat{w})}
\frac{1}{n}\mathrm{Tr}\frac{\partial^2 g(\hat{w})}{\partial w^2}I_X(w^*)^{-1}+o_p\bigg(\frac{1}{n}\bigg)\bigg\}\nonumber\\
&= -\ln g(\hat{w}) -\frac{1}{2g(\hat{w})}\frac{1}{n}\mathrm{Tr}\frac{\partial^2 g(\hat{w})}{\partial w^2}I_X(w^*)^{-1}+o_p\bigg(\frac{1}{n}\bigg).\label{eq:last_term}
\end{align}
Using the Taylor expansion at $w^*$, we obtain
\begin{align*}
\frac{1}{\xi(z|\hat{w})} =  \frac{1}{\xi(z|w^*)}+o_p\bigg(\frac{1}{n}\bigg).
\end{align*}
Considering this equation and the order of $r(\hat{w})$ with Eq. \ref{eq:MLE_order},
we average Eq. \ref{eq:last_term}:
\begin{align*}
&-E_zE_n\bigg[ \ln \int g(w)\mathcal{N}(w|\hat{w},\{nI_X(w^*)\}^{-1})dw\bigg]\\
&=-E_zE_n\big[\ln \xi(z|\hat{w})\varphi(\hat{w})\big]\\
&\;\;\;\;-\frac{1}{2n}E_n\bigg[ \frac{1}{\varphi(\hat{w})e^{r(\hat{w})}}
\mathrm{Tr}\frac{\partial^2}{\partial w^2}\bigg\{E_z\bigg[\frac{\xi(z|\hat{w})}{\xi(z|w^*)}\bigg]\varphi(\hat{w})e^{r(\hat{w})}\bigg\}I_X(w^*)^{-1}\bigg]
+o\bigg(\frac{1}{n}\bigg)\\
&=-E_zE_n\big[\ln \xi(z|\hat{w})\varphi(\hat{w})\big]
-\frac{1}{2n}E_n\bigg[ \frac{1}{\varphi(\hat{w})}
\mathrm{Tr}\frac{\partial^2 \varphi(\hat{w})}{\partial w^2}I_X(w^*)^{-1}\bigg]
+o\bigg(\frac{1}{n}\bigg),
\end{align*}
where the last expression is based on $E_z[\frac{\xi(z|w)}{\xi(z|w^*)}]=1$ for any $w$.
Replacing this expression with the last term of Eq. \ref{eq:Fxi_mid},
we obtain Eq. \ref{eq:Fxi}.
\subsection{Proof of Theorem \ref{th:DII}}
%
In the case of $\xi(z|w)=p(x_j,y_j|w)$,
\begin{align*}
F_1(n) =& E_zE_{n-1}\bigg[-\ln \int \xi(z|w)\prod_{i\neq j}^n p(x_i|w)\varphi(w)dw\bigg]\\
=& F_{\xi}(n-1).
\end{align*}
In the case of $\xi(z|w)=p(x_j|w)$,
\begin{align*}
F_2(n) =& E_zE_{n-1}\bigg[-\ln \int \xi(z|w)\prod_{i\neq j}^n p(x_i|w)\varphi(w)dw\bigg]\\
=& F_{\xi}(n-1).
\end{align*}

Based on Lemma \ref{lem:Fxi}, $F_1(n)$ and $F_2(n)$ can be written as
\begin{align*}
F_1(n) =& -E_n\bigg[\sum_{i\ne j}^n \ln p(x_i|\hat{w}_{n-1})\bigg]+\frac{1}{2}\ln \frac{1}{2\pi} \det\{(n-1)I_X(w^*)\}\\
&-E_n[\ln p(x_j,y_j|\hat{w}_{n-1})\varphi(\hat{w}_{n-1})]\\
&-\frac{1}{2(n-1)}\mathrm{Tr}E_n\bigg[\frac{1}{\varphi(\hat{w}_{n-1})}
\frac{\partial^2 \varphi(\hat{w}_{n-1})}{\partial w^2}\bigg]I_X(w^*)^{-1}
+o\bigg(\frac{1}{n}\bigg),\\
F_2(n) =& -E_n\bigg[\sum_{i\ne j}^n \ln p(x_i|\hat{w}_{n-1})\bigg]+\frac{1}{2}\ln \frac{1}{2\pi} \det\{(n-1)I_X(w^*)\}\\
&-E_n[\ln p(x_j|\hat{w}_{n-1})\varphi(\hat{w}_{n-1})]\\
&-\frac{1}{2(n-1)}\mathrm{Tr}E_n\bigg[\frac{1}{\varphi(\hat{w}_{n-1})}
\frac{\partial^2 \varphi(\hat{w}_{n-1})}{\partial w^2}\bigg]I_X(w^*)^{-1}
+o\bigg(\frac{1}{n}\bigg),
\end{align*}
where we used the asymptotic behavior of $\hat{w}_{n-1}$ and the order shown in Eq. \ref{eq:MLE_n-1_order}.

Using these forms and the relation of Eq. \ref{eq:DIIFF},
we obtain
\begin{align*}
D^{\mathrm{Bayes}}_{\mathrm{II}}(n) =& \frac{1}{n} \sum_{j=1}^n E_n\bigg[
\ln\frac{q(y_j|x_j)}{p(y_j|x_j,\hat{w}_{n-1})}\bigg]+o\bigg(\frac{1}{n}\bigg).
\end{align*}
According to the definition of the error for Type III,
\begin{align*}
E_n\bigg[\ln\frac{q(y_j|x_j)}{p(y_j|x_j,\hat{w}_{n-1})}\bigg]
=D^{\mathrm{ML}}_{\mathrm{III}}(n-1)+o\bigg(\frac{1}{n}\bigg),
\end{align*}
which is independent of $j$.
Thus,
\begin{align*}
D^{\mathrm{Bayes}}_{\mathrm{II}}(n) =& D^{\mathrm{ML}}_{\mathrm{III}}(n-1) + o\bigg(\frac{1}{n}\bigg).
\end{align*}
Using Theorem \ref{th:prev_results}, we can derive the following form:
\begin{align*}
D^{\mathrm{Bayes}}_{\mathrm{II}}(n) =& \frac{\mathrm{Tr}I_{Y|X}(w^*)I_X(w^*)^{-1}}{2(n-1)} + o\bigg(\frac{1}{n}\bigg),
\end{align*}
which proves Theorem \ref{th:DII}.
\subsection{Proof of Theorem \ref{th:DIII}}
\label{sec:proof_DIII}
%
In the case of $\xi(z|w)=p(y_{n+1}|x_{n+1},w)$,
\begin{align*}
F_3(n) =& E_zE_n\bigg[ - \ln \int \xi(z|w)\prod_{i=1}^n p(x_i|w)\varphi(w)dw\bigg]\\
=& F_{\xi}(n).
\end{align*}
Based on Lemma \ref{lem:Fxi}, $F_3(n)$ can be rewritten as
\begin{align*}
F_3(n) =& E_{n+1} \bigg[ -\sum_{i=1}^n \ln p(x_i|\hat{w})
- \frac{1}{2} \ln 2\pi\det\{nI_X(w^*)\}^{-1} \\
&-\ln p(y_{n+1}|x_{n+1},\hat{w})\varphi(\hat{w})
- \frac{1}{2n}\mathrm{Tr}\frac{1}{\varphi(\hat{w})}\frac{\partial^2 \varphi(\hat{w})}{\partial w^2}I_X(w^*)^{-1} \bigg] 
+o\bigg(\frac{1}{n}\bigg).
\end{align*}

Let us assume that the function $\xi$ is a constant $\xi(z|w)=1$.
Then, $F_2(n)$ has another expression;
\begin{align*}
F_2(n) =& E_zE_n\bigg[ -\ln \int 1\cdot \prod_{i=1}^n p(x_i|w)\varphi(w)dw\bigg]\\
=& F_{\xi}(n).
\end{align*}
It is easily confirmed that
Lemma \ref{lem:Fxi} holds in this case,
and $F_2(n)$ has the following form;
\begin{align*}
F_2(n) =& E_n \bigg[ -\sum_{i=1}^n \ln p(x_i|\hat{w})
- \frac{1}{2} \ln 2\pi\det\{nI_X(w^*)\}^{-1} \\
&-\ln \varphi(\hat{w})
- \frac{1}{2n}\mathrm{Tr}\frac{1}{\varphi(\hat{w})}\frac{\partial^2 \varphi(\hat{w})}{\partial w^2}I_X(w^*)^{-1} \bigg] 
+o\bigg(\frac{1}{n}\bigg).
\end{align*}

Using these forms and the relation of Eq. \ref{eq:DIIIFF},
we obtain
\begin{align*}
D^{\mathrm{Bayes}}_{\mathrm{III}}(n) =& E_{n+1}\bigg[\ln \frac{q(y_{n+1}|x_{n+1})}{p(y_{n+1}|x_{n+1},\hat{w})}\bigg] +o\bigg(\frac{1}{n}\bigg)\\
=& D^{\mathrm{ML}}_{\mathrm{III}}(n)+o\bigg(\frac{1}{n}\bigg).
\end{align*}
According to Theorem \ref{th:prev_results},
\begin{align*}
D^{\mathrm{Bayes}}_{\mathrm{III}}(n) = &\frac{\mathrm{Tr}I_{Y|X}(w^*)I_X(w^*)^{-1}}{2n} + o\bigg(\frac{1}{n}\bigg),
\end{align*}
which proves Theorem \ref{th:DIII}.
\section{Discussion}
\label{sec:dis}
In the previous section,
we found that the accuracy of the Bayes estimation was asymptotically equivalent to that of the maximum-likelihood estimation for Types II and III.
In this section, we investigate the mathematical reason 
why the Bayes estimation is advantageous for Type I.

In Section \ref{sec:variants}, Types II and III are extended to multivariable estimations,
and their asymptotic errors are introduced.
The results indicate that the Bayes method is again more accurate.
In Section \ref{sec:multi-prediction},
we compare single-variable and multivariable predictions,
and we find that the Bayes estimation is advantageous
not only when estimating latent variables but also when estimating observable variables.
In Section \ref{sec:analysis_mve},
we formally decompose the error functions of the multivariable estimations
and elucidate the difference between the Bayes and maximum-likelihood methods.

\subsection{Other Estimations of Multiple Latent Variables}
\label{sec:variants}
Let us consider the variants of Types II and III,
in which there are multiple estimation targets.
We consider a positive constant $\alpha$,
where $\alpha n$ is an integer.
We will use the following notation for the data:
\begin{align*}
X_1 =& \{x_1,\dots,x_{\alpha n}\},\\
Y_1 =& \{y_1,\dots,y_{\alpha n}\},\\
X_2 =& \{x_{n+1},\dots,x_{n+\alpha n}\},\\
Y_2 =& \{y_{n+1},\dots,y_{n+\alpha n}\}.
\end{align*}
\begin{definition}[Type II$'$]
Assume that $0<\alpha<1$.
Let $X^n$ be the observable data, and let $Y_1$ be the estimation targets.
The maximum-likelihood estimation is given by
\begin{align*}
p(Y_1|X^n) = \prod_{i=1}^{\alpha n} p(y_i|x_i,\hat{w})
= \prod_{i=1}^{\alpha n}\frac{p(x_i,y_i|\hat{w})}{p(x_i|\hat{w})},
\end{align*}
and the Bayes estimation is given by
\begin{align*}
p(Y_1|X^n) =& \frac{\int \prod_{j=1}^{\alpha n} p(x_j,y_j|w)\prod_{i=\alpha n+1}^n p(x_i|w)\varphi(w;\eta)dw}
{\int \prod_{i=1}^n p(x_i|w)\varphi(w;\eta)dw}.
\end{align*}
\end{definition}
In Type II$'$, the estimation is on the joint probability of $Y_1$,
where $Y^n\setminus Y_1$ is marginalized out.
Note that $X^n\setminus X_1=\{x_{\alpha n+1},\dots,x_n\}$
is used for both estimations,
since $\hat{w}$ is based on $X^n$ in the maximum-likelihood method
and the numerator and the denominator include the factor $\prod_{i=\alpha n+1}^n p(x_i|w)$
in the Bayes method.
Type II$'$ lies between Types I and II;
it is equivalent to Type I for $\alpha =1$
and formally converges to Type II as $\alpha\rightarrow 1/n$.
\begin{definition}[Type III$'$]
Let $X^n$ and $X_2$ be the observable data, and let $Y_2$ be the estimation targets.
The maximum-likelihood estimation is given by
\begin{align*}
p(Y_2|X^n,X_2) = \prod_{i=n+1}^{n+\alpha n} p(y_i|x_i,\hat{w})
= \prod_{i=n+1}^{n+\alpha n}\frac{p(x_i,y_i|\hat{w})}{p(x_i|\hat{w})},
\end{align*}
and the Bayes estimation is given by
\begin{align*}
p(Y_2|X^n,X_2) =& \int \prod_{i=n+1}^{n+\alpha n}\frac{p(x_i,y_i|w)}{p(x_i|w)}p(w|X^n)dw.
\end{align*}
\end{definition}
In Type III$'$, the estimation is on the joint probability of $Y_2$.
Type III$'$ formally converges to Type III as $\alpha\rightarrow 1/n$.

\begin{figure}[t]
\centering
\includegraphics[angle=-90,width=0.7\columnwidth]{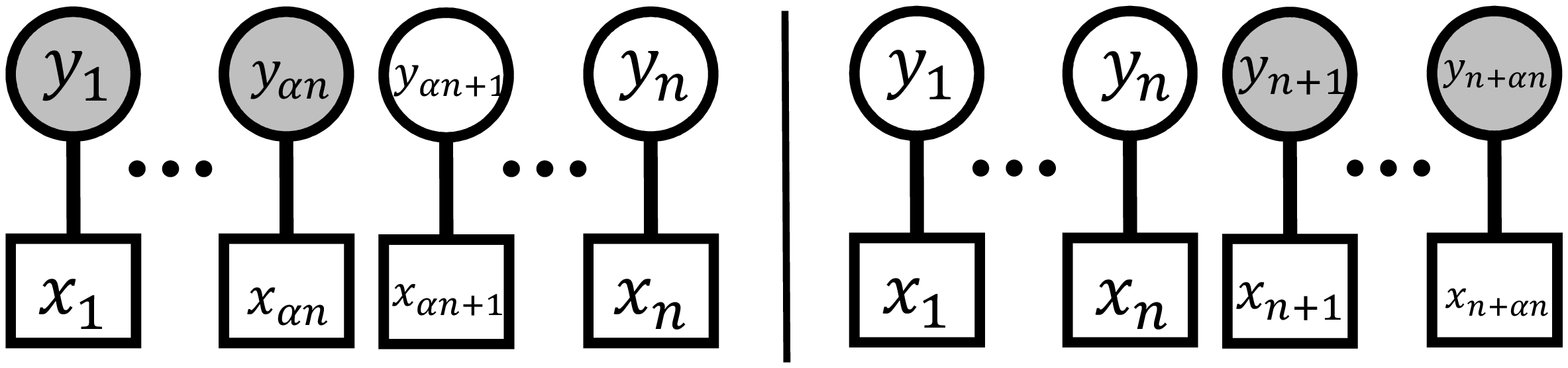}
\caption{Variants of Types II and III.}
\label{fig:Types_v}
\end{figure}
Figure \ref{fig:Types_v} shows these types;
the left panel shows Type II$'$, which is the multitarget estimation of Type II,
and the right panel shows Type III$'$, which is the multitarget estimation of Type III.

The error functions of Type II$'$ and III$'$ are defined by
\begin{align*}
D_{\mathrm{II'}}(n) =& \frac{1}{\alpha n}E_{X^n}\bigg[\sum_{Y_1}q(Y_1|X^n)\ln\frac{q(Y_1|X^n)}{p(Y_1|X^n)}\bigg],\\
D_{\mathrm{III'}}(n) =& \frac{1}{\alpha n}E_{X^n,X_2}\bigg[\sum_{Y_2}q(Y_2|X_2)\ln\frac{q(Y_2|X_2)}{p(Y_2|X_2,X^n)}\bigg],
\end{align*}
respectively.
In the maximum-likelihood estimation,
according to the definitions,
$D_{\mathrm{II'}}(n)=D_{\mathrm{II}}(n)$ and $D_{\mathrm{III'}}(n)=D_{\mathrm{III}}(n)$.
By comparing $D_{\mathrm{II'}}(n)$ and $D_{\mathrm{III'}}(n)$
with $D_{\mathrm{II}}(n)$ and $D_{\mathrm{III}}(n)$, respectively,
we can clarify whether the Bayes method is advantageous
for multivariable estimations.

Let us define a mixture of the Fisher information matrices:
\begin{align*}
K_{XY}(w)=\alpha I_{XY}(w)+(1-\alpha)I_X(w).
\end{align*}
In a previous study \citep{Yamazaki14a}, we proved the following lemmas.
\begin{lemma}
\label{lem:type2d}
In the Bayes estimation for Type II$'$, the error function has the following asymptotic expansion:
\begin{align*}
D^{\mathrm{Bayes}}_{\mathrm{II'}}(n) =& \frac{1}{2\alpha n}\ln\det[K_{XY}(w^*)I_X(w^*)^{-1}] +o\bigg(\frac{1}{n}\bigg).
\end{align*}
\end{lemma}
\begin{lemma}
\label{lem:type3d}
In the Bayes estimation for Type III$'$, the error function has the following asymptotic expansion:
\begin{align*}
D^{\mathrm{Bayes}}_{\mathrm{III'}}(n) =& \frac{1}{2\alpha n}\ln\det[K_{XY}(w^*)I_X(w^*)^{-1}]+o\bigg(\frac{1}{n}\bigg).
\end{align*}
\end{lemma}
These lemmas show the following relations, based on the leading terms:
\begin{align*}
D^{\mathrm{Bayes}}_{\mathrm{II'}}(n) &< D^{\mathrm{ML}}_{\mathrm{II'}}(n)
=D^{\mathrm{ML}}_{\mathrm{II}}(n),\\
D^{\mathrm{Bayes}}_{\mathrm{III'}}(n) &< D^{\mathrm{ML}}_{\mathrm{III'}}(n)
=D^{\mathrm{ML}}_{\mathrm{III}}(n).
\end{align*}
By comparing these relations with Corollary \ref{cor:mag_rel},
we see that the Bayes method is advantageous
when there are multiple estimation targets:
\begin{align*}
D^{\mathrm{Bayes}}_{\mathrm{II'}}(n) <& D^{\mathrm{Bayes}}_{\mathrm{II}}(n),\\
D^{\mathrm{Bayes}}_{\mathrm{III'}}(n) <& D^{\mathrm{Bayes}}_{\mathrm{III}}(n).
\end{align*}

\subsection{Estimation of Multiple Observable Variables}
\label{sec:multi-prediction}
In the previous subsection,
it was proved that the Bayes method was advantageous 
for all multivariable estimations of latent variables.
Let us consider the following two cases for estimating observable variables.
\begin{definition}[Single-target prediction]
Let $X^n$ be the observable data, and let $x_{n+1}$ be the estimation target.
The maximum-likelihood estimation is given by
\begin{align*}
p(x_{n+1}|X^n) =& p(x_{n+1}|\hat{w}),
\end{align*}
and the Bayes estimation is given by
\begin{align*}
p(x_{n+1}|X^n) =& \int p(x_{n+1}|w)p(w|X^n)dw.
\end{align*}
\end{definition}
\begin{definition}[Multiple-target prediction]
Let $X^n$ be the observable data, and let $X_2$ be the estimation target.
The maximum-likelihood estimation is given by
\begin{align*}
p(X_2|X^n) =& \prod_{i=n+1}^{n+\alpha n} p(x_i|\hat{w}),
\end{align*}
and the Bayes estimation is given by
\begin{align*}
p(X_2|X^n) =& \int \prod_{i=n+1}^{n+\alpha n} p(x_i|w)p(w|X^n)dw.
\end{align*}
\end{definition}
\begin{figure}[t]
\centering
\includegraphics[angle=-90,width=0.7\columnwidth]{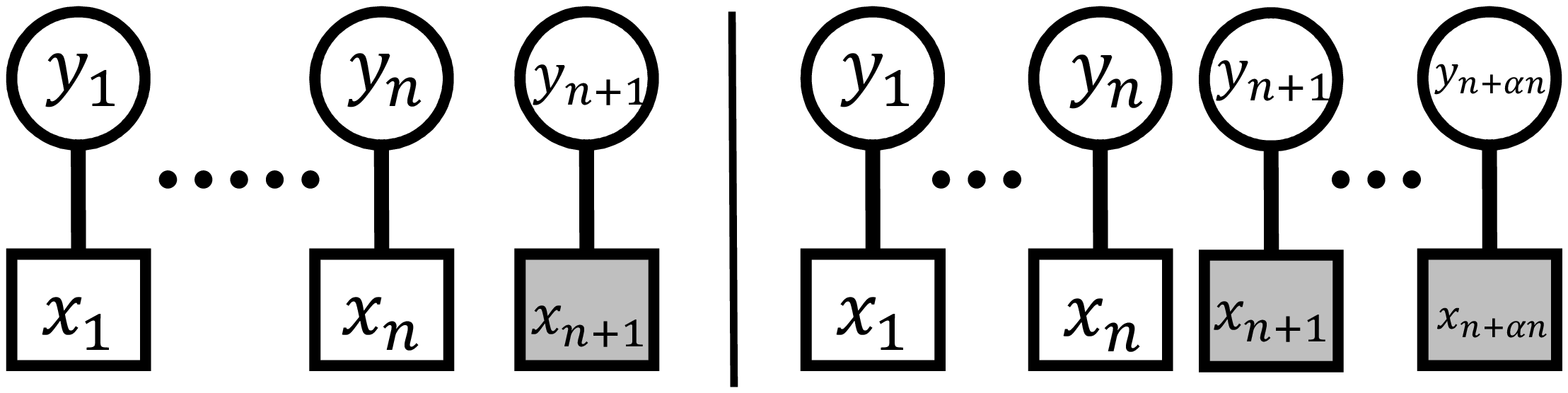}
\caption{Predictions of a single observable variable (the left panel)
and of multiple variables (the right panel).}
\label{fig:Pred_v}
\end{figure}
Figure \ref{fig:Pred_v} shows these predictions;
the left and right panels show the predictions for a single target and for multiple targets,
respectively.

The error functions for the single-target prediction (STP) and multiple-target prediction (MTP)
are defined by
\begin{align*}
D_{\mathrm{STP}}(n) =& E_{n+1}\bigg[\ln \frac{q(x_{n+1})}{p(x_{n+1}|X^n)}\bigg],\\
D_{\mathrm{MTP}}(n) =& \frac{1}{\alpha n}E_{n+\alpha n}\bigg[ \ln \frac{q(X_2)}{p(X_2|X^n)}\bigg],
\end{align*}
respectively.

The following lemma shows that we obtain a smaller error with the Bayes method
not only when estimating the latent variables 
but also when estimating the observable variables.
\begin{lemma}
\label{lem:pred_errors}
The error functions in the predictions have the following asymptotic expansions:
\begin{align*}
D^{\mathrm{ML}}_{\mathrm{STP}}(n) =& \frac{d}{2n} +o\bigg(\frac{1}{n}\bigg),\\
D^{\mathrm{ML}}_{\mathrm{MTP}}(n) =& D^{\mathrm{ML}}_{\mathrm{STP}}(n),\\
D^{\mathrm{Bayes}}_{\mathrm{STP}}(n) =& D^{\mathrm{ML}}_{\mathrm{STP}}(n) + o\bigg(\frac{1}{n}\bigg),\\
D^{\mathrm{Bayes}}_{\mathrm{MTP}}(n) =& \frac{\ln (1+\alpha)}{\alpha}\frac{d}{2n} +o\bigg(\frac{1}{n}\bigg),
\end{align*}
where $d$ is the dimension of the parameter.
\end{lemma}
The proofs are given in the Appendix.
We can now obtain the following relations, based on the leading terms:
\begin{align*}
D^{\mathrm{Bayes}}_{\mathrm{MTP}}(n) < D^{\mathrm{ML}}_{\mathrm{MTP}}(n)
= D^{\mathrm{ML}}_{\mathrm{STP}}(n) = D^{\mathrm{Bayes}}_{\mathrm{STP}}(n).
\end{align*}
Again, we see that the Bayes estimation is more accurate in the multiple-target case,
and its accuracy is equivalent to that of the maximum-likelihood estimation in the single-target case.

\subsection{Analysis of the Advantage in Multivariable Estimations}
\label{sec:analysis_mve}
In the Bayesian multivariable estimations,
the estimated distributions are defined by the integrals of the parameter,
where all target variables are also used for their estimation.
For example, the Bayes estimation of the multiple-target prediction is given by
\begin{align*}
p(X_2|X^n) =& \int \prod_{i=n+1}^{n+\alpha n} p(x_i|w)p(w|X^n)dw\\
=& \frac{\int \prod_{i=1}^{n+\alpha n}p(x_i|w)\varphi(w;\eta)dw}{\int \prod_{i=1}^n p(x_i|w)\varphi(w;\eta)dw},
\end{align*}
where the numerator includes the likelihood of $X^n\cup X_2=\{x_1,\dots, x_{n+\alpha n}\}$.
On the other hand, the maximum-likelihood estimation is based on the likelihood of $X^n$.
This implies that the dependent way of the Bayes estimation
will be more accurate than repetition of the single-variable estimation.

We can mathematically explain this advantage as follows.
In the prediction problem, the MTP error is formally expressed as
\begin{align*}
D_{\mathrm{MTP}}(n) = \frac{1}{\alpha n}E_{n+\alpha n}\bigg[ \ln q(X_2) &- \ln p(X_2|X^n) \bigg]\\
= \frac{1}{\alpha n}E_{n+\alpha n}\bigg[ \ln q(X_2) &- \ln p(x_{n+1}|X_2\setminus x_{n+1},X^n)\\
 &-\ln p(X_2\setminus x_{n+1}|X^n) \bigg]\\
= \frac{1}{\alpha n}E_{n+\alpha n}\bigg[ \ln q(X_2) &- \ln p(x_{n+1}|X_2\setminus x_{n+1},X^n)\\
&- \ln p(x_{n+2}|X_2\setminus\{x_{n+1},x_{n+2}\},X^n)\\
&-\ln p(X_2\setminus\{x_{n+1},x_{n+2}\}|X^n)\bigg]\\
= \frac{1}{\alpha n}E_{n+\alpha n}\bigg[ \sum_{i=1}^{\alpha n} \ln q(x_{n+i}) &- \ln p(x_{n+1}|X_2\setminus x_{n+1},X^n)\\
&- \ln p(x_{n+2}|X_2\setminus\{x_{n+1},x_{n+2}\},X^n)\\
&\dots - \ln p(x_{n+\alpha n -1}|x_{n+\alpha n},X^n)\\
&-\ln p(x_{n+\alpha n}|X^n)\bigg].
\end{align*}
Then,
\begin{align*}
D_{\mathrm{MTP}}(n) =& \frac{1}{\alpha n} \sum_{i=1}^{\alpha n} D_{\mathrm{MTP},i}(n),\\
D_{\mathrm{MTP},i}(n) =& E_{n+\alpha n}\bigg[ \ln \frac{q(x_{n+i})}{p(x_{n+i}|X_2\setminus\{x_{n+1},\dots,x_{n+i}\},X^n)}\bigg].
\end{align*}
Because the maximum-likelihood estimation determines $\hat{w}$ from $X^n$,
\begin{align*}
D^{\mathrm{ML}}_{\mathrm{MTP},i}(n) =& E_{n+\alpha n}\bigg[ \ln \frac{q(x_{n+i})}{p(x_{n+i}|\hat{w})}\bigg]
= D^{\mathrm{ML}}_{\mathrm{STP}}(n),
\end{align*}
which means that
\begin{align*}
D^{\mathrm{ML}}_{\mathrm{MTP}}(n) =& \frac{1}{\alpha n} \sum_{i=1}^{\alpha n} D^{\mathrm{ML}}_{\mathrm{STP}}(n) 
= D^{\mathrm{ML}}_{\mathrm{STP}}(n).
\end{align*}
Comparing this with the maximum-likelihood estimation $p(x_{n+i}|X^n)=p(x_{n+i}|\hat{w})$,
we find that the Bayes estimation $p(x_{n+i}|X_2\setminus\{x_{n+1},\dots,x_{n+i}\},X^n)$
uses the additional data set $X_2\setminus\{x_{n+1},\dots,x_{n+i}\}$,
which results in a more accurate prediction.

Now, we consider the estimation of latent variables.
Let us define the following notation:
\begin{align*}
Y^n_i =& Y^n\setminus \{y_i,\dots,y_n\}=\{y_1,\dots,y_{i-1}\},\\
Y_{1,i} =& Y_1\setminus \{y_i,\dots,y_{\alpha n}\}=\{y_1,\dots,y_{i-1}\},\\
Y_{2,i} =& Y_2\setminus\{y_{n+i},\dots,y_{n+\alpha n}\}=\{y_{n+1},\dots,y_{n+i-1}\}.
\end{align*}
For example, the estimated probability of Type I can be written as
\begin{align*}
p(Y^n|X^n) =& p(y_n|Y^n_n,X^n)p(Y^n_n|X^n)\\
=&p(y_n|Y^n_n,X^n)p(y_{n-1}|Y^n_{n-1},X^n)p(Y^n_{n-1}|X^n)\\
=& \prod_{i=1}^n p(y_i|Y^n_i,X^n).
\end{align*}
In the same way,
\begin{align*}
p(Y_1|X^n) =& \prod_{i=1}^{\alpha n}p(y_i|Y_{1,i},X^n),\\
p(Y_2|X^n,X_2) =& \prod_{i=1}^{\alpha n}p(y_{n+i}|Y_{2,i},X_2,X^n),
\end{align*}
for Type II$'$ and III$'$, respectively.
Then, the error functions can be rewritten as
\begin{align*}
D_{\mathrm{I}}(n) =& \frac{1}{n}\sum_{i=1}^n D_{\mathrm{I},i}(n),\\
D_{\mathrm{II'}}(n) =& \frac{1}{\alpha n}\sum_{i=1}^{\alpha n} D_{\mathrm{II'},i}(n),\\
D_{\mathrm{III'}}(n) =& \frac{1}{\alpha n}\sum_{i=1}^{\alpha n} D_{\mathrm{III'},i}(n),
\end{align*}
where
\begin{align*}
D_{\mathrm{I},i}(n) =& E_n\bigg[ \ln \frac{q(y_i|x_i)}{p(y_i|Y^n_i,X^n)} \bigg],\\
D_{\mathrm{II'},i}(n) =& E_n\bigg[ \ln \frac{q(y_i|x_i)}{p(y_i|Y_{1,i},X^n)} \bigg],\\
D_{\mathrm{III'},i}(n) =& E_{n+\alpha n}\bigg[\ln \frac{q(y_{n+i}|x_{n+i})}{p(y_{n+i}|Y_{2,i},X_2,X^n)}\bigg],
\end{align*}
respectively.
Note that, in these formal product forms,
a target $y_i$ is estimated based on the results of other targets;
for example, Type I has the probability $p(y_i|Y^n_i,X^n)$,
where $y_i$ depends on the results of $Y^n_i=\{y_1,\dots,y_{i-1}\}$.
However, in the maximum-likelihood method, the estimated probabilities are expressed as
\begin{align*}
p(y_i|Y^n_i,X^n) =& p(y_i|\hat{w}) = p(y_i|X^n),\\
p(y_i|Y_{1,i},X^n) =& p(y_i|\hat{w}) = p(y_i|X^n),\\
p(y_{n+i}|Y_{2,i},X_2,X^n) =& p(y_{n+i}|x_{n+i},\hat{w}) = p(y_{n+i}|x_{n+i},X^n),
\end{align*}
respectively,
where additional data, such as $Y^n_i$, $Y_{1,i}$, and $Y_{2,i}$, is ignored.
It can easily be found that for $i\ge 2$,
\begin{align*}
D^{\mathrm{ML}}_{\mathrm{I},i}(n) =& E_n\bigg[ \ln \frac{q(y_i|x_i)}{p(y_i|X^n)} \bigg]\\
&> E_n\bigg[ \ln \frac{q(y_i|x_i)}{p(y_i|Y^n_i,X^n)} \bigg] = D^{\mathrm{Bayes}}_{\mathrm{I},i}(n),\\
D^{\mathrm{ML}}_{\mathrm{II'},i}(n) =& E_n\bigg[ \ln \frac{q(y_i|x_i)}{p(y_i|X^n)} \bigg]\\
&> E_n\bigg[ \ln \frac{q(y_i|x_i)}{p(y_i|Y_{1,i},X^n)} \bigg] = D^{\mathrm{Bayes}}_{\mathrm{II'},i}(n),\\
D^{\mathrm{ML}}_{\mathrm{III'},i}(n) =& E_{n+\alpha n}\bigg[\ln \frac{q(y_{n+i}|x_{n+i})}{p(y_{n+i}|X_2,X^n)}\bigg]\\
&> E_{n+\alpha n}\bigg[\ln \frac{q(y_{n+i}|x_{n+i})}{p(y_{n+i}|Y_{2,i},X_2,X^n)}\bigg] 
= D^{\mathrm{Bayes}}_{\mathrm{III'},i}(n),
\end{align*}
which shows that the error of the maximum-likelihood method is larger than that of the Bayes method.

Let us consider the single-variable estimations from the perspective of this additional data.
In the multivariable estimation, the Bayes method has an advantage, 
because the error functions defined by the Kullback-Leibler divergence
are decomposed into terms such as $D_{\mathrm{I},i}(n)$, $D_{\mathrm{II'},i}(n)$,
and $D_{\mathrm{III'},i}(n)$, which express the error on each $y_i$. Thus, the use of additional data, such as $Y^n_i$, $Y_{1,i}$, and $Y_{2,i}$,
improves the accuracy.
Note that these data points are also the estimation targets in other terms.
On the other hand,
the single-variable estimations do not have any other targets,
and thus the error function does not decompose and the Bayes method does not have an advantage.
Using Theorems \ref{th:DII} and \ref{th:DIII},
we quantitatively confirmed that the asymptotic accuracies of the Bayes and maximum-likelihood methods were equal.
\section{Conclusion}
The present paper derived the asymptotic accuracy of the Bayes latent-variable estimation
for Types II and III, which are both single-variable estimations.
The results indicate that the accuracy of the Bayes method 
is equivalent to that of the maximum-likelihood method. 
This clarifies that the Bayes method is only advantageous
for multivariable estimations, such as Types I, II$'$, and III$'$.
\section*{Acknowledgments}
This research was partially supported by the CASIO Science Promotion Foundation
and KAKENHI 23500172.
%
\section*{Appendix}
\subsection*{Proof of Lemma \ref{lem:pred_errors}}
Since the first equation is a well-known result and is shown in the literature \citep{Akaike,Watanabe09:book}, 
we omit the proof.

The second equation is derived from the definitions of the error functions:
\begin{align*}
D^{\mathrm{ML}}_{\mathrm{MTP}}(n) =& 
\frac{1}{\alpha n} E_{n+\alpha n}\bigg[ \sum_{i=n+1}^{n+\alpha n}\ln \frac{q(x_i)}{p(x_i|\hat{w})}\bigg]\\
=& \frac{1}{\alpha n}\sum_{i=n+1}^{n+\alpha n} E_{n+\alpha n}\bigg[\ln \frac{q(x_i)}{p(x_i|\hat{w})}\bigg]\\
=& \frac{1}{\alpha n}\sum_{i=n+1}^{n+\alpha n} D^{\mathrm{ML}}_{\mathrm{STP}}(n)\\
=& D^{\mathrm{ML}}_{\mathrm{STP}}(n).
\end{align*}

Using the form of $F_2(n)$ shown in Section \ref{sec:proof_DIII}, we obtain 
\begin{align*}
D^{\mathrm{Bayes}}_{\mathrm{STP}}(n) =& E_{n+1}\big[ \ln q(x_{n+1}) + F_2(n+1) -F_2(n)\big]\\
=& E_{n+1}\bigg[\sum_{i=1}^{n+1} \ln q(x_i) + F_2(n+1)\bigg] - E_n\bigg[ \sum_{i=1}^n \ln q(x_i) +F_2(n)\bigg]\\
=& \frac{d}{2}\ln(n+1) - \frac{d}{2}\ln n + o\bigg(\frac{1}{n}\bigg)\\
=& \frac{d}{2n} +o\bigg(\frac{1}{n}\bigg),
\end{align*}
which proves the third equation.

Based on the same form of $F_2(n)$, the last equation is derived as follows:
\begin{align*}
D^{\mathrm{Bayes}}_{\mathrm{MTP}}(n)
=& \frac{1}{\alpha n}\bigg\{E_{n+\alpha n}\bigg[ \sum_{i=1}^{n+\alpha n} \ln q(x_i) +F_2(n+\alpha n)\bigg]\\
&-E_n\bigg[ \sum_{i=1}^n \ln q(x_i) + F_2(n)\bigg] \bigg\}\\
=& \frac{1}{\alpha n}\bigg\{ \frac{d}{2}\ln(n+\alpha n) -\frac{d}{2}\ln n\bigg\} +o\bigg(\frac{1}{n}\bigg)\\
=& \frac{\ln(1+\alpha)}{\alpha}\frac{d}{2n} +o\bigg(\frac{1}{n}\bigg).
\end{align*}
\vskip 0.2in
\bibliography{LearningTheory}
\bibliographystyle{natbib}
\end{document}